\documentclass{article}

\usepackage[preprint]{neurips_2022}

\usepackage{graphicx}
\usepackage[utf8]{inputenc} 
\usepackage[T1]{fontenc}    
\usepackage{hyperref}       
\usepackage{url}            
\usepackage{booktabs}       
\usepackage{amsfonts}       
\usepackage{nicefrac}       
\usepackage{microtype}      
\usepackage{xcolor}         
\usepackage{amsmath}
\usepackage{comment}
\title{Introducing One Sided Margin Loss for Solving Classification Problems in Deep Networks}

\author{
   Ali Karimi \\
   University of Tehran \\
   Tehran, Iran \\
   aliiikarimi@ut.ac.ir \\
   
   \And
   Zahra Mousavi Kouzehkanan \\
   University of Tehran \\
   Tehran, Iran \\
   z\_mousavi@ut.ac.ir \\
   \AND
   Reshad Hosseini \\
   University of Tehran \\
   Tehran, Iran \\
   reshad.hosseini@ut.ac.ir \\
   \And
   Hadi Asheri \\
   University of Tehran \\
   Tehran, Iran \\
   hadi.asheri@ut.ac.ir \\
}

\begin{document}

\maketitle

\begin{abstract}

This paper introduces a new loss function, OSM (One-Sided Margin), to solve maximum-margin classification problems effectively. Unlike the hinge loss, in OSM the margin is explicitly determined with corresponding hyperparameters and then the classification problem is solved. In experiments, we observe that using OSM loss leads to faster training speeds and better accuracies than binary and categorical cross-entropy in several commonly used deep models for classification and optical character recognition problems.

OSM has consistently shown better classification accuracies over cross-entropy and hinge losses for small to large neural networks. it has also led to a more efficient training procedure. We achieved state-of-the-art accuracies for small networks on several benchmark datasets of CIFAR10(98.82\%), CIFAR100(91.56\%), Flowers(98.04\%), Stanford Cars(93.91\%) with considerable improvenemts over other loss functions. Moreover, the accuracies are rather better than cross entropy and hinge loss for large networks. Therefore, we strongly believe that OSM is a powerful alternative to hinge and cross-entropy losses to train deep neural networks on classification tasks.

\end{abstract}

\section{Introduction}

  In the past decade, neural networks has made a significant progress for differnet classification problems. Since AlexNet (\cite{krizhevsky2012imagenet}), there has been a lot of competition to obtain more accurate neural networks. Deeper and more complex networks (\cite{simonyan2014very}), better optimization algorithms (\cite{reddi2019convergence}),improved activation functions (\cite{rakhlin2018land}) and better loss functions  (\cite{bonyadi2019optimal}) hase been proposed.
  
  Research on the loss function is a hotspot in this area because an appropriate loss function allows for achieving a more accurate model and might lead to a more efficient training procedure (\cite{han2017novel,bosman2020visualising}). A well-known loss function, especially in the literature of support vector machines (SVMs), is the hinge loss (\cite{lee2015deeply}) which is commonly used for maximum-margin classification. The loss function of an SVM classifier is a linear combination of the empirical error and the margin. Both the empirical error and the margin are functions of the weight vector. During the training procedure, this vector must be chosen to optimize both of them simultaneously. The problem with this configuration is that the width of the achieved margin is not guaranteed. This is a common problem with already known maximum margin classifiers.
  
\begin{figure}[ht]
  \centering
  \label{fig:OSM_planes}
\includegraphics[width=0.5\linewidth]{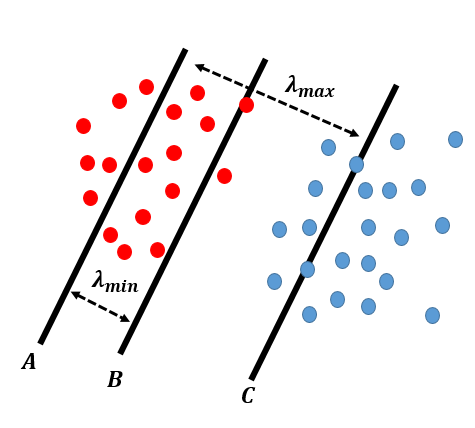}
  \caption{The margin hyperplanes in OSM loss. The osm loss has a fixed and predefined margin and tries to locate positive data between the hyperplanes A and B  and push the negative ones beyond the hyperplane C.}
\end{figure}

  In this paper, we propose a novel loss function, called OSM, to guarantee the width of the obtained classification margin. In fact, OSM allows us to first determine the width of the margin and then adjust the parameters such that the classification error is minimized. The binary OSM loss for data point $(x,y)$ is defined in the following equation:

\begin{equation}
l_{\text{Binary OSM}} = y\max\left( s - \lambda_{\min},0\  \right) + \lambda\max\left( \lambda_{\max} - s,0 \right)(1 - y) + \alpha\max( - s,0)y
\label{eq.osmBinary}
\end{equation}
    
  where  $s = \mathbf{w}^T\mathbf{x}+b $ is the score, $y \in \{0,1\}$ is the label, $\alpha$ is the Lagrange multiplier, $\lambda$, $\lambda_{\min }$, and $\lambda_{\max }$ are the hyperparameters and $\lambda_{\max } - \lambda_{\min }$ determines the margin width. A visual explanation of the OSM loss has been shown in Figure \ref{fig:OSM_planes}.

  The major contributions of this paper can be summarized
as follows:

\begin{itemize}
\item We propose OSM loss function to achieve effective maximum margin classifiers for image classification and optical character recognition (OCR) tasks.
\item We applied OSM loss on commonly used convolutional neural networks for different image classification datasets.
\item We combined OSM with connectionist temporal classification (CTC) loss to achieve a good loss function for OCR problems.
\item We provide an extensive set of experiments on several datasets to verify the effectiveness and efficiency of the OSM loss in comparison to hinge and cross entropy loss functions.

\end{itemize}

\section{Related Works}

  A loss function is a measure of how good a prediction model does in terms of being able to predict the expected outcome. In fact, we convert the learning problem into an optimization problem, define a loss function and then adjust the model parameters to minimize the defined loss function. To improve the accuracies on classification tasks, previous works have proposed many loss functions including mean square error (\cite{chang2006modified}), cross-entropy (CE) (\cite{kohonen1988statistical}), hinge \cite{lee2015deeply}, kullback leibler divergence (\cite{csiszar1975divergence}), gussian radial basis function (RBF) \cite{zadeh2018deep} and negative log-likelihood (\cite{lopez2015hybrid}), perceptrons  \cite{minsky1969introduction} and logarithmic function (\cite{gasso2019logistic}).

  Before the rise of deep neural networks, hinge loss and its variations were the most commonly used loss functions for classification problems(\cite{hui2020evaluation}).
  Differnt variants of hinge loss were proposed including ramp (\cite{collobert2006trading}), pinball(\cite{huang2013support}), rescaled hinge (\cite{xu2017robust}), squared hinge (\cite{lee2015deeply}), twin hinge (\cite{huang2018twin}) and truncated pinball (\cite{shen2017support}). Linear, polynomial, gaussian RBF and hiperbolic tangent SVMs are among the most effective type of max-margin classifiers using hinge loss (\cite{cervantes2020comprehensive}).

  To train deep neural networks with hinge loss, usually a linear SVM layer is used (\cite{wang2022comprehensive}). Although cross-entropy is known as the most popular loss function, linear SVM has shown comparable performances for image classification with deep neural networks (\cite{tang2013deep,goodfellow2016deep}).

\section{Proposed Method}

In this section, we explain the core idea behind the OSM loss function. Lets assume that \(x\) is
input data and \(y \in \{ 0,1,2,..,C - 1\}\) is the corresponding class label.
Consider the function \(f\) maps x to an output vector
\(s = f(x) \in R^{C}\) which represents the scores of different classes.
In the context of deep neural networks, $f$ is the network and $s$ is the final score vector. For simplicity, lets first consider the 2-class
problem, i.e. \(y \in \left\{ 0,1 \right\}\) and \(s \in R\). The one sided margin loss function, for a single input data, is computed as: 

\begin{equation}
\label{eq.osm1}
l_{\text {Binary OSM }}=y \max \left(s-\lambda_{\min }, 0\right)+\lambda \max \left(\lambda_{\max }-s, 0\right)(1-y) \quad \text { s.t. } y s \geq 0
\end{equation}

where, \(\lambda_{\max} > \lambda_{\min} \geq 0\ \) and
\(\lambda > 0\) are the hyperparameters.

When $y$ is $0$, the condition
\(\ ys \geq 0\ \) is satisfied and when the output score is greater than or equal to
 \(\lambda_{\max}\), the loss function becomes
zero and a score less than \(\lambda_{\max}\)
causes a nonzero loss. When $y$ is $1$, the condition \(ys \geq 0\) means that
$s$ must be non-negative and the loss function also tries to get $s$
closer to \(\lambda_{\min}\). In other words, a positive $s$ less than or equal to \(\lambda_{\min}\) makes the loss function zero.

If we ignore the constraint of the OSM loss, it becomes very
similar to the hinge loss used in SVMs. Unlike
the SVM loss function, which utilizes the regularization term
\(\left\| w \right\|\) to enforce the margin in the final feature space,
our loss considers a predefined fixed margin. 
As explained in Figure \ref{fig:OSM_planes}, the OSM loss tries to locate
the positive data between the hyperplanes $A$ (0) and $B$ (\(\lambda_{\min}\)) and push the negative
ones beyond the hyperplane $C$ (\(\lambda_{\max}\)). It is straightforward to add the contraint term to the loss function as a Lagrange multiplier. 

In the OSM loss, the scale of input data isn't important and no specific normalization is needed. This desired characterstic is achived because the norm of the weight vector is not used in the loss, unlike the SVMs. For example, if the input feature is scaled using matrix $A$ the final hyperplane is also scaled  similarly.

The multiclass OSM loss is defind in \eqref{eq.osm}:
\begin{equation}
\label{eq.osm}
l_{\text{OSM}} = \max\left( s_{y} - \lambda_{\min},0 \right) + \alpha\max\left( - s_{y},0 \right) + \lambda\sum_{j \neq y}^{}{\max\left( \lambda_{\max} - s_{j},0 \right)}
\end{equation}
where  \(s_{y}\) is the score of the true class.

Until now, we have defined the hard version of the OSM loss function with a difficult optimization process due to using the $max(.)$ function. We also propose the soft version of the OSM loss by replacing $\max(.,0)$ with $\log\bigl(1+\exp(.)\bigr)$, as depicted in \eqref{osm.main}:

\begin{equation}
\label{osm.main}
l_{\text {soft-OSM }}=\log \left(1+e^{s_{y}-\lambda_{\min }}\right)+\alpha \log \left(1+e^{-s_{y}}\right)+\lambda \sum_{j \neq y} \log \left(1+e^{\lambda_{\max }-s_{j}}\right)
\end{equation}

According to \eqref{osm.r}, we can assume that OSM loss maximizes the
logarithm of the unnormalized probabilities of classes. So, the normalized
probability of assigning label $k$ to input data $x$ can easily be computed
by using $\log{p\left( y = k \middle| x \right)}=-l_{soft - OSM}+c,\quad k \in \left\{ 0,1,2,\ \ldots,\ C - 1\right\}$, wehre $c$ is an arbitrary normalization constant. With this configuration of loss function, it is also possible to consider a rejection class. In other words, when the probabilities of all classes are small, we assume that the data belongs to an extra class called rejection class ($C$). The probability of rejection is defined as follows:

\begin{equation}
\label{osm.r}
\log p(y=C \mid \mathbf{x})=\lambda \sum_{j} \log \left(1+e^{\lambda_{\max }-s_{j}}\right)
\end{equation}

Although in classification tasks, there is no need to use the above defined probabilities, because the class with minimum score $s$ is chosen as the prediction, it might be very useful for many other tasks.  
For example, in the OCR tasks, unnormalized probabilities can be calculated before feeding
the scores into the CTC loss function and also the consider the blank
character as the rejection class.

\section{Experimental Results}

In this section, we conduct different experiments with OSM loss for image classification and OCR tasks. Different deep neural networks are examined over various image classification datasets in Section \ref{imageclassification}. We also analyze the combination of OSM and CTC losses for OCR tasks in Section \ref{ocrsec}.

\subsection{Image Classification}
\label{imageclassification}

This section contains the comparison of the proposed OSM loss with cross-entropy and hinge loss functions on different networks over well-known image datasets.

\subsubsection{Implementation Details}
\begin{figure}[ht]
  \centering
\includegraphics[width=0.4\linewidth]{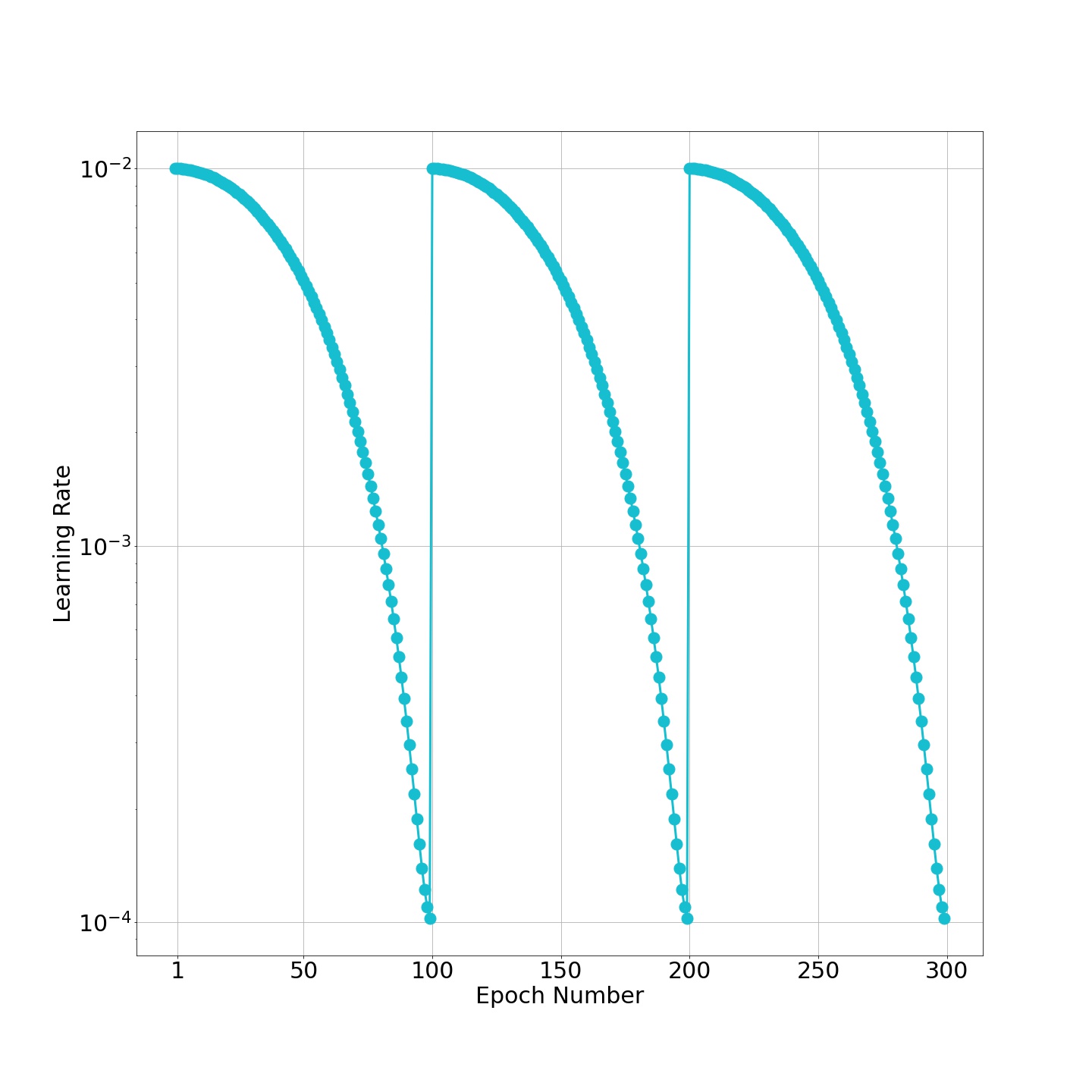}
  \caption{Trend of learning rate during neural network training}
  \label{learning}
\end{figure}

To optimize all examined loss functions, we utilize stochastic gradient descent (SGD) with the configurations shown in Table \ref{table1}.
The initial value of the learning rate is 0.01 and it is iteratively updated according to the cosine annealing with warmup scheduler (\cite{loshchilov2016sgdr}) that restarts the learning rate 100 epochs. The learning rate plot is shown in Fig \ref{learning}.

As shown in Table \ref{table1}, some of the parameters are common among cross-entropy, hinge, and OSM loss functions and hyperparameters $\alpha$, $\lambda$, $\lambda_{\min }$, and $\lambda_{\max }$, are specific to the OSM loss and $\alpha$ is specific to hinge loss. 
We have reported the average output of 3 runs for each experiment.

\begin{table}[ht]
  \caption{Hyperparameter of neural networks in image classification experiments}
  \centering
  \label{table1}
  \begin{tabular}{c|c|c}
    \toprule
    \cmidrule(r){1-2}
    Loss function     & Parameter     & Value  \\
    \midrule
     & Epoch & 300    \\
    & Optimizer & SGD+Momentum       \\
    Hinge, CE and OSM & Momentum  & 0.9     \\
         & Regularization & L2     \\
         & Weight Decay & 0.0005     \\
         & Batch size & 32     \\
        & Initial Learning Rate & 0.01     \\
    
    \midrule
    Hinge     & $\alpha$      & 1  \\
    
    \midrule
         & $\alpha$       & 0.1  \\
    OSM     & $\lambda$      & 1  \\
         & $\lambda_{\max }$       & 600  \\
         & $\lambda_{\min }$       & 100  \\
    \bottomrule
  \end{tabular}
\end{table}

\subsubsection{Datasets \& Architectures}

In this experiment, we have examined 5 state-of-the-art convolutional architectures including  VGG16 (\cite{simonyan2014very}), EfficientNetB0 (\cite{tan2019efficientnet}), EfficientNetV2-S, EfficientNetV2-M and EfficientNetV2-L (\cite{tan2021efficientnetv2}) on 4 benchmark datasets including CIFAR10 (\cite{krizhevsky2009learning}), CIFAR100 (\cite{krizhevsky2009learning}), Stanford Cars (\cite{krause2013collecting}) and Flowers (\cite{nilsback2008automated}). 
Detained information about each dataset is shown in Table \ref{table2}. 
Table \ref{table3} presents details about different convolutional architectures. 

\begin{table}[ht]
\caption{Image classification datasets}
\centering
\label{table2}
\begin{tabular}{c|ccc}
\toprule
Dataset & Train Size & Test Size & \#Classes \\
\hline CIFAR-10 \cite{krizhevsky2009learning} & 50,000 & 10,000 & 10 \\
CIFAR-100 \cite{krizhevsky2009learning} & 50,000 & 10,000 & 100 \\
Stanford Cars \cite{krause2013collecting} & 8,144 & 8,041 & 196 \\
Flowers \cite{nilsback2008automated} & 2,040 & 6,149 & 102 \\
\bottomrule
\end{tabular}
\end{table}

\begin{table}[ht]
\caption{Image classification architectures}
\centering
\label{table3}
\begin{tabular}{c|c}
\toprule
Model & Parameters  \\
\hline VGG16  (\cite{simonyan2014very}) & 138.4M	 \\
EfficientNetB0 (\cite{tan2019efficientnet}) & 5.3M  \\
EfficientNetV2-S (\cite{tan2021efficientnetv2}) & 21.6M	 \\
EfficientNetV2-M (\cite{tan2021efficientnetv2}) & 54.4M	 \\
EfficientNetV2-L (\cite{tan2021efficientnetv2}) & 119.0M \\
\bottomrule
\end{tabular}
\end{table}

\subsubsection{Evaluation Metric}

Different metrics such as F1-Score, accuracy, precision and recall be used to compare the models obtained by every loss function. In this experiment, we use accuracy as a common evaluation criterion:

\begin{equation}
\label{acc}
Accuracy =    \frac{True Positive  + True Negative}{Total}
\end{equation}

\subsubsection{Results}
To determine the appropriate values for hyperparameters of OSM loss, we trained a VGG16 network with different configurations on CIFAR10 dataset as shown in Table \ref{table4}. According to the obtained results, we set $\lambda$ to $1$,  $\alpha$ to $0.1$, $\lambda_{\max }$ to $600$ and $\lambda_{\min}$ to 100.

\begin{table}[h!]
  \caption{Comparison of the effect of hyperparameters on the OSM loss function}
  \label{table4}
  \centering
  \begin{tabular}{c|c|c|c|c}
    \toprule
    \cmidrule(r){1-2}
    $\alpha$     & $\lambda$     & $\lambda_{\max }$  & $\lambda_{\min }$ & Accuracy  \\
    \midrule
     0.01 & 1  & 600 & 100 & 90.65  \\
     0.1 & 1  & 600 & 100 & 92.77 \\
     1 & 1  & 600 & 100 & 92.14 \\
     5 & 1  & 600 & 100 & 91.65  \\
     10 & 1  & 600 & 100 & 91.19 \\
    \midrule
     0.1 & 0.01  & 600 & 100 & 91.08  \\
     0.1 & 0.1  & 600 & 100 & 92.43 \\
     0.1 & 1  & 600 & 100 & 92.77  \\
     0.1 & 5  & 600 & 100 & 91.82  \\
      0.1 & 10  & 600 & 100 & 90.94  \\
    \midrule
    
     0.1 & 1  & 1000 & 0 & 91.75  \\
     0.1 & 1  & 700 & 0 & 92.09 \\
     0.1 & 1  & 500 & 0 & 92.68  \\
     0.1 & 1  & 300 & 0 & 92.13  \\
     0.1 & 1  & 100 & 0 & 91.89 \\
     0.1 & 1  & 900 & 400 & 92.49 \\
     0.1 & 1  & 800 & 300 & 92.55 \\
     0.1 & 1  & 700 & 200 & 92.61  \\
     0.1 & 1  & 600 & 100 & 92.77  \\
     0.1 & 1  & 500 & 0 & 92.63 \\
    \bottomrule
  \end{tabular}
\end{table}

\begin{figure}[ht]
  \centering
\includegraphics[width=0.5\linewidth]{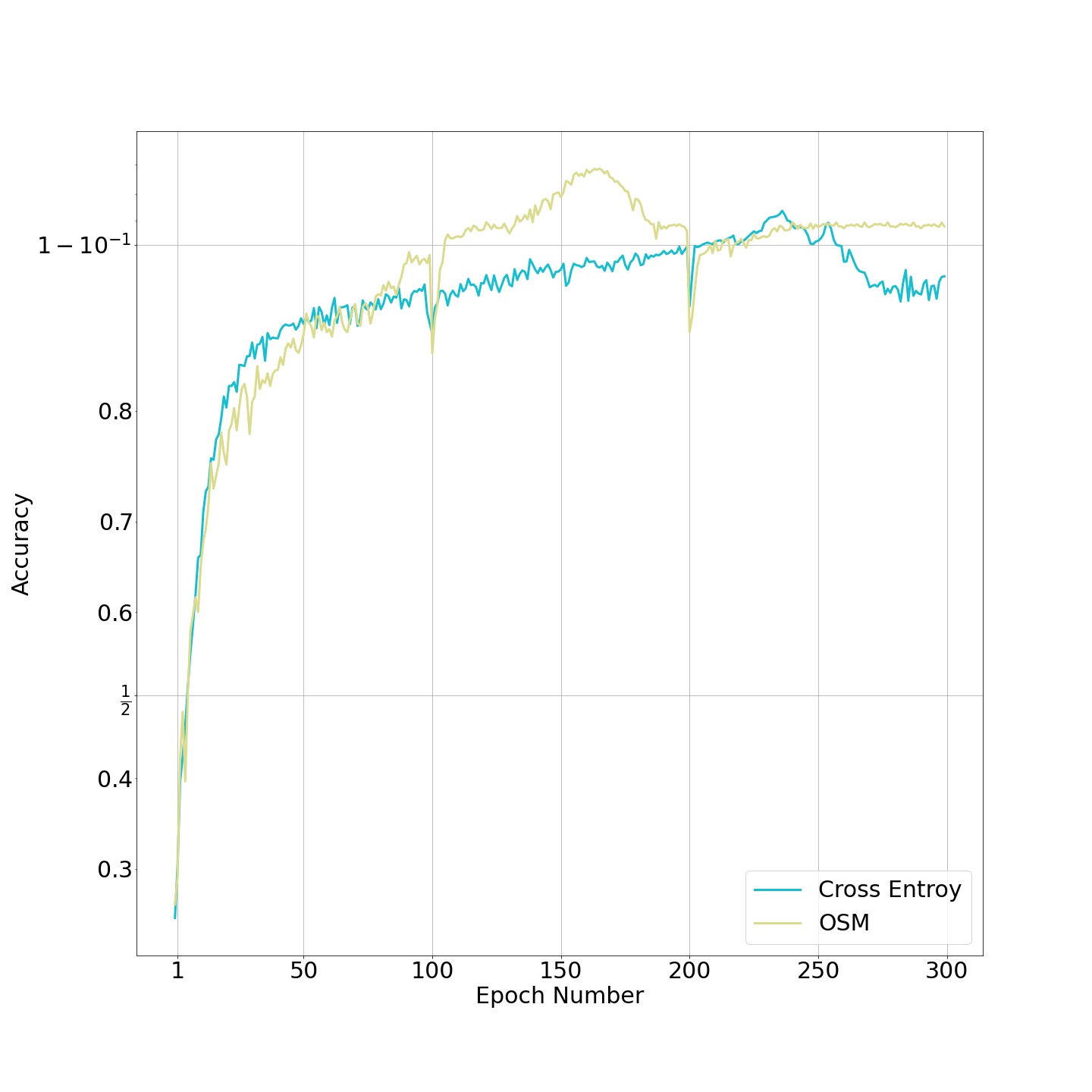}
  \caption{Comparison accuracy on CIFAR10, training on VGG16}
  
  \label{figaccuracy}
\end{figure}

As shown in Table \ref{table5} and Fig \ref{figaccuracy}, the training procedure of OSM converges faster and achieves better accuracy in comparison to other loss functions. For example, for EfficientNetB0 trained on CIFAR-10, OSM achieves an accuracy about 1.01\%  better than cross-entropy and 1.25\% better than hinge loss.
Some interesting observations about OSM can be summarized as follows:

\begin{itemize}
\item In general, in all models and datasets, the accuracy of the OSM loss function is higher than others, for example, an improvement about 0.45\% on CIFAR10. In other words, OSM loss consistently results in classifiers better than or equal to the ones obtained by other loss functions.
\item For smaller datasets, there is a bigger gap between OSM and the other loss functions. For example, for the EfficientNetB0 network, the accuracy on the CIFAR10 dataset with 10 classes has increased about 1.01\%, while it is about 0.6\% for CIFAR100 with 100 classes. The improvement is a bit smaller (0.41) for the Stanford Cars dataset with 196 classes.
\item For small networks, OSM depicts a bigger improvement in accuracy than the obtained improvement for large networks. For example, on the CIFAR10 dataset, OSM has shown an improvement about 1.01\% while its improvement for the EfficientNetV2-L network 0.15\%.
\item 
Along with better accuracies, OSM has demonstrated a faster training procedure compared to cross-entropy. For the VGG16 model on the CIFAR10 dataset, the OSM loss function has achived the accuracy 92.53\% in epoch 139  while cross-entropy
has achieved the same accuracy after 237 epochs.
\item As explained in the previous section, all hyperparameters are adjusted in a general fashin, i.e. from the training of VGG16 on CIFAR10. It is obvious that adjusting the hyperparameters for each dataset will result to better performance.
\item It can be also observed that cross-entropy loss is always better than hinge loss for all models and datasets.
\end{itemize}

\begin{table}[ht]
  \caption{Comparison between networks and datasets (EffNet is the abbreviation of EfficientNet).}
  \label{table5}
  \centering
\makebox[\textwidth][c]{
  \begin{tabular}{c|c|c|c|c|c|c}
    \toprule
    Dataset     & Loss Function     & VGG16  &   EffNetB0 & EffNetV2-S & EffNetV2-M & EffNetV2-L \\
    \midrule
     & OSM & 92.77   & 98.53 & 98.82  & 99.01 & 99.05   \\
    CIFAR10 & CE & 92.53   & 97.52 & 98.19 & 98.78 & 98.90        \\
     & Hinge  & 92.38 & 97.28 & 98.10  & 98.02 & 98.12     \\
     & Improvement  & 0.24  & 1.01 & 0.63 & 0.23 & 0.15     \\

    \midrule
     & OSM & 98.59 & 99.54 & 99.60 & 99.66 & 99.78    \\
    CIFAR10  & Binary CE & 98.43 & 99.10 & 99.23 & 99.50 & 99.71        \\
     (Plane,Truck)& Hinge  & 98.12 & 98.72 & 98.97 & 99.03 & 99.21      \\
     & Improvement  & 0.16 & 0.44 & 0.37 & 0.16 & 0.07      \\

    \midrule
     & OSM & 71.38  & 88.41 & 91.56 & 92.25 & 92.20    \\
    CIFAR100 & CE & 71.29  & 87.81 & 91.17 & 92.08 & 92.12       \\
     & Hinge  & 70.88  & 87.38 & 90.88 & 91.65 & 91.78      \\
     & Improvement  & 0.09  & 0.60 & 0.39 & 0.17 & 0.08      \\

    \midrule
   
     & OSM & 87.36 &  97.13 & 98.04  & 98.51 & 98.69   \\
    Flowers & CE & 87.14  & 96.62 & 97.61 & 98.35 & 98.64      \\
     & Hinge  & 86.43  & 96.19 & 97.03 & 97.92 & 98.02      \\
     & Improvement  & 0.22  & 0.51 & 0.43 & 0.16 & 0.05      \\
     
    \midrule
     & OSM & 88.02  & 91.06 & 93.91 & 94.50 & 94.99   \\
    Stanford Cars & CE & 87.96 & 90.65 & 93.53  & 94.41 & 94.95       \\
     & Hinge  & 87.19  & 90.21 & 92.92  & 94.13 & 94.67     \\
     & Improvement  & 0.06  & 0.41 & 0.38  & 0.09 & 0.04    \\

    \bottomrule
  \end{tabular}
}
\end{table}

\subsection{Optical Character Recognition}
\label{ocrsec}
In this section, the common CTC loss will be compared with the combined OSM-CTC loss in which, instead of scores, normalized probabilities obtained from OSM are fed into CTC. Both loss functions are examined on the license plate number recognition problem. During this experiment, the blank character is assumed as the rejection class.

\subsubsection{Implementation Details}

Two optical character recognition models are implemented using TensorFlow version 1.15.0. The initial value of the learning rate is 0.001. When using OSM layer before CTC loss, like the classification problem, cosine annealing with warmup has been used to change the learning rate in different epochs. For common CTC loss, the cosine annealing scheduler didn't show a good performance but the exponential decay scheduler has shown a very better performance. The values of common CTC and OSM hyper-parameters and OSM-specific hyper-parameters are shown in Table \ref{table6}.

\begin{table}[ht]
  \caption{Hyperparameter of neural networks in optical character recognition experiments}
  \label{table6}
  \centering
  \begin{tabular}{c|c|c}
    \toprule
    \cmidrule(r){1-2}
    Loss function     & Parameter     & Value  \\
    \midrule
     & Epoch & 300    \\
    CTC and CTC-OSM & Optimizer & Adam       \\
     & Batch size & 60     \\
     & Initial Learning Rate & 0.001     \\
    \midrule
         & $\alpha$       & 1  \\
    OSM     & $\lambda$      & 1  \\
         & $\lambda_{\max }$       & 600  \\
         & $\lambda_{\min }$       & 100  \\
    \bottomrule
  \end{tabular}
\end{table}

\subsubsection{Datasets \& Architectures}

In this task, we used the LPRNet architecture (\cite{zherzdev2018lprnet}), which is a real-time license plate recognition network and is trained end-to-end. The CCPD (Chinese City Parking Dataset) (\cite{xu2018towards}) dataset is also used to train and evaluate the LPRNet model. The number of training and validation data are 100000 and 99996, respectively. We considered both the basic LPRNet and its scaled-down version with fewer parameters. In the scaled-down version, the number of channels is reduced from 64 to 16, and the channel increasing coefficient is halved.

%

\subsubsection{Evaluation Metric}

In order to evaluate the optical character recognition quantitatively,
the ratio of correctly recognized plates to the total number of license plates
is computed (Equation. \eqref{ac}). The greedy decoder was used to generate the
output sequence of characters. The greedy decoder is very
straightforward because the letter with the highest probability
is selected at each time step.

\begin{equation}
\text {Acc}_\mathrm{OCR}=\frac{\text { The number of correctly recognized license plates }}{\text { Total number of license plates }}
\label{ac}
\end{equation}

\subsubsection{Results}

Table \ref{table8} shows the accuracies obtained by the OSM-CTC and CTC loss functions. Although OSM loss function achieves better accuracies for both models, its improvement is more significant (1.8\%) for the smaller model.

\begin{table}[ht]
  \caption{Comparasion between OSM-CTC and CTC}
  \label{table8}
  \centering
  \begin{tabular}{c|c|c}
    \toprule
    Model - Loss function     & CTC     & OSM  \\
    \midrule
    scaled-down LPRNet model     & 0.958       & 0.976  \\
    LPRNet model     & 0.99       & 0.995  \\
    \bottomrule
  \end{tabular}
\end{table}

\section{Conclusion}

We proposed a novel loss function called OSM that creates a fixed margin for the decision boundaries between different classes. It is very straightforward and intuitive to set the hyperparameters of OSM loss. Extensive experiments show that OSM achieves better results with different network architectures on benchmark datasets compared to other well-known loss functions, in both image classification and OCR tasks. According to experiments, OSM loss demonstrates a bigger improvement for smaller networks. Also, the probabilities produced by OSM improved the performance of the OCR task especially when the model is scaled down. 

\subsection{Extentions}
Since OSM loss leads to big improvements over hinge and cross entropy loss functions for small architectures, it is definitley the best choice to run deep models on low resource and embedded devices. OSM loss might also be helpful for other vision tasks like object detection. It can also be applied on audio and text classification tasks.

\bibliographystyle{unsrtnat}
\bibliography{main.bib}

\begin{thebibliography}{33}
\providecommand{\natexlab}[1]{#1}
\providecommand{\url}[1]{\texttt{#1}}
\expandafter\ifx\csname urlstyle\endcsname\relax
  \providecommand{\doi}[1]{doi: #1}\else
  \providecommand{\doi}{doi: \begingroup \urlstyle{rm}\Url}\fi

\bibitem[Krizhevsky et~al.(2012)Krizhevsky, Sutskever, and
  Hinton]{krizhevsky2012imagenet}
Alex Krizhevsky, Ilya Sutskever, and Geoffrey~E Hinton.
\newblock Imagenet classification with deep convolutional neural networks.
\newblock \emph{Advances in neural information processing systems}, 25, 2012.

\bibitem[Simonyan and Zisserman(2014)]{simonyan2014very}
Karen Simonyan and Andrew Zisserman.
\newblock Very deep convolutional networks for large-scale image recognition.
\newblock \emph{arXiv preprint arXiv:1409.1556}, 2014.

\bibitem[Reddi et~al.(2019)Reddi, Kale, and Kumar]{reddi2019convergence}
Sashank~J Reddi, Satyen Kale, and Sanjiv Kumar.
\newblock On the convergence of adam and beyond.
\newblock \emph{arXiv preprint arXiv:1904.09237}, 2019.

\bibitem[Rakhlin et~al.(2018)Rakhlin, Davydow, and Nikolenko]{rakhlin2018land}
Alexander Rakhlin, Alex Davydow, and Sergey Nikolenko.
\newblock Land cover classification from satellite imagery with u-net and
  lov{\'a}sz-softmax loss.
\newblock In \emph{Proceedings of the IEEE Conference on Computer Vision and
  Pattern Recognition Workshops}, pages 262--266, 2018.

\bibitem[Bonyadi and Reutens(2019)]{bonyadi2019optimal}
Mohammad~Reza Bonyadi and David~C Reutens.
\newblock Optimal-margin evolutionary classifier.
\newblock \emph{IEEE Transactions on Evolutionary Computation}, 23\penalty0
  (5):\penalty0 885--898, 2019.

\bibitem[Han and Wu(2017)]{han2017novel}
Bin Han and Yiquan Wu.
\newblock A novel active contour model based on modified symmetric cross
  entropy for remote sensing river image segmentation.
\newblock \emph{Pattern Recognition}, 67:\penalty0 396--409, 2017.

\bibitem[Bosman et~al.(2020)Bosman, Engelbrecht, and
  Helbig]{bosman2020visualising}
Anna~Sergeevna Bosman, Andries Engelbrecht, and Mard{\'e} Helbig.
\newblock Visualising basins of attraction for the cross-entropy and the
  squared error neural network loss functions.
\newblock \emph{Neurocomputing}, 400:\penalty0 113--136, 2020.

\bibitem[Lee et~al.(2015)Lee, Xie, Gallagher, Zhang, and Tu]{lee2015deeply}
Chen-Yu Lee, Saining Xie, Patrick Gallagher, Zhengyou Zhang, and Zhuowen Tu.
\newblock Deeply-supervised nets.
\newblock In \emph{Artificial intelligence and statistics}, pages 562--570.
  PMLR, 2015.

\bibitem[Chang and Carin(2006)]{chang2006modified}
Shaorong Chang and Lawrence Carin.
\newblock A modified spiht algorithm for image coding with a joint mse and
  classification distortion measure.
\newblock \emph{IEEE transactions on image processing}, 15\penalty0
  (3):\penalty0 713--725, 2006.

\bibitem[Kohonen et~al.(1988)Kohonen, Barna, and
  Chrisley]{kohonen1988statistical}
Teuvo Kohonen, Gy{\"o}rgy Barna, and Ronald~L Chrisley.
\newblock Statistical pattern recognition with neural networks: benchmarking
  studies.
\newblock In \emph{ICNN}, volume~1, page~61, 1988.

\bibitem[Csisz{\'a}r(1975)]{csiszar1975divergence}
Imre Csisz{\'a}r.
\newblock I-divergence geometry of probability distributions and minimization
  problems.
\newblock \emph{The annals of probability}, pages 146--158, 1975.

\bibitem[Zadeh et~al.(2018)Zadeh, Hosseini, and Sra]{zadeh2018deep}
Pourya~Habib Zadeh, Reshad Hosseini, and Suvrit Sra.
\newblock Deep-rbf networks revisited: Robust classification with rejection.
\newblock \emph{arXiv preprint arXiv:1812.03190}, 2018.

\bibitem[Lopez-Garcia et~al.(2015)Lopez-Garcia, Onieva, Osaba, Masegosa, and
  Perallos]{lopez2015hybrid}
Pedro Lopez-Garcia, Enrique Onieva, Eneko Osaba, Antonio~D Masegosa, and Asier
  Perallos.
\newblock A hybrid method for short-term traffic congestion forecasting using
  genetic algorithms and cross entropy.
\newblock \emph{IEEE Transactions on Intelligent Transportation Systems},
  17\penalty0 (2):\penalty0 557--569, 2015.

\bibitem[Minsky and Papert(1969)]{minsky1969introduction}
Marvin Minsky and Seymour Papert.
\newblock An introduction to computational geometry.
\newblock \emph{Cambridge tiass., HIT}, 479:\penalty0 480, 1969.

\bibitem[Gasso(2019)]{gasso2019logistic}
Gilles Gasso.
\newblock Logistic regression, 2019.

\bibitem[Hui and Belkin(2020)]{hui2020evaluation}
Like Hui and Mikhail Belkin.
\newblock Evaluation of neural architectures trained with square loss vs
  cross-entropy in classification tasks.
\newblock \emph{arXiv preprint arXiv:2006.07322}, 2020.

\bibitem[Collobert et~al.(2006)Collobert, Sinz, Weston, and
  Bottou]{collobert2006trading}
Ronan Collobert, Fabian Sinz, Jason Weston, and L{\'e}on Bottou.
\newblock Trading convexity for scalability.
\newblock In \emph{Proceedings of the 23rd international conference on Machine
  learning}, pages 201--208, 2006.

\bibitem[Huang et~al.(2013)Huang, Shi, and Suykens]{huang2013support}
Xiaolin Huang, Lei Shi, and Johan~AK Suykens.
\newblock Support vector machine classifier with pinball loss.
\newblock \emph{IEEE transactions on pattern analysis and machine
  intelligence}, 36\penalty0 (5):\penalty0 984--997, 2013.

\bibitem[Xu et~al.(2017)Xu, Cao, Hu, and Principe]{xu2017robust}
Guibiao Xu, Zheng Cao, Bao-Gang Hu, and Jose~C Principe.
\newblock Robust support vector machines based on the rescaled hinge loss
  function.
\newblock \emph{Pattern Recognition}, 63:\penalty0 139--148, 2017.

\bibitem[Huang et~al.(2018)Huang, Wei, and Zhou]{huang2018twin}
Huajuan Huang, Xiuxi Wei, and Yongquan Zhou.
\newblock Twin support vector machines: A survey.
\newblock \emph{Neurocomputing}, 300:\penalty0 34--43, 2018.

\bibitem[Shen et~al.(2017)Shen, Niu, Qi, and Tian]{shen2017support}
Xin Shen, Lingfeng Niu, Zhiquan Qi, and Yingjie Tian.
\newblock Support vector machine classifier with truncated pinball loss.
\newblock \emph{Pattern Recognition}, 68:\penalty0 199--210, 2017.

\bibitem[Cervantes et~al.(2020)Cervantes, Garcia-Lamont,
  Rodr{\'\i}guez-Mazahua, and Lopez]{cervantes2020comprehensive}
Jair Cervantes, Farid Garcia-Lamont, Lisbeth Rodr{\'\i}guez-Mazahua, and
  Asdrubal Lopez.
\newblock A comprehensive survey on support vector machine classification:
  Applications, challenges and trends.
\newblock \emph{Neurocomputing}, 408:\penalty0 189--215, 2020.

\bibitem[Wang et~al.(2022)Wang, Ma, Zhao, and Tian]{wang2022comprehensive}
Qi~Wang, Yue Ma, Kun Zhao, and Yingjie Tian.
\newblock A comprehensive survey of loss functions in machine learning.
\newblock \emph{Annals of Data Science}, 9\penalty0 (2):\penalty0 187--212,
  2022.

\bibitem[Tang(2013)]{tang2013deep}
Yichuan Tang.
\newblock Deep learning using linear support vector machines.
\newblock \emph{arXiv preprint arXiv:1306.0239}, 2013.

\bibitem[Goodfellow et~al.(2016)Goodfellow, Bengio, and
  Courville]{goodfellow2016deep}
Ian Goodfellow, Yoshua Bengio, and Aaron Courville.
\newblock \emph{Deep learning}.
\newblock MIT press, 2016.

\bibitem[Loshchilov and Hutter(2016)]{loshchilov2016sgdr}
Ilya Loshchilov and Frank Hutter.
\newblock Sgdr: Stochastic gradient descent with warm restarts.
\newblock \emph{arXiv preprint arXiv:1608.03983}, 2016.

\bibitem[Tan and Le(2019)]{tan2019efficientnet}
Mingxing Tan and Quoc Le.
\newblock Efficientnet: Rethinking model scaling for convolutional neural
  networks.
\newblock In \emph{International conference on machine learning}, pages
  6105--6114. PMLR, 2019.

\bibitem[Tan and Le(2021)]{tan2021efficientnetv2}
Mingxing Tan and Quoc Le.
\newblock Efficientnetv2: Smaller models and faster training.
\newblock In \emph{International Conference on Machine Learning}, pages
  10096--10106. PMLR, 2021.

\bibitem[Krizhevsky et~al.(2009)Krizhevsky, Hinton,
  et~al.]{krizhevsky2009learning}
Alex Krizhevsky, Geoffrey Hinton, et~al.
\newblock Learning multiple layers of features from tiny images.
\newblock 2009.

\bibitem[Krause et~al.(2013)Krause, Deng, Stark, and
  Fei-Fei]{krause2013collecting}
Jonathan Krause, Jia Deng, Michael Stark, and Li~Fei-Fei.
\newblock Collecting a large-scale dataset of fine-grained cars.
\newblock 2013.

\bibitem[Nilsback and Zisserman(2008)]{nilsback2008automated}
Maria-Elena Nilsback and Andrew Zisserman.
\newblock Automated flower classification over a large number of classes.
\newblock In \emph{2008 Sixth Indian Conference on Computer Vision, Graphics \&
  Image Processing}, pages 722--729. IEEE, 2008.

\bibitem[Zherzdev and Gruzdev(2018)]{zherzdev2018lprnet}
Sergey Zherzdev and Alexey Gruzdev.
\newblock Lprnet: License plate recognition via deep neural networks.
\newblock \emph{arXiv preprint arXiv:1806.10447}, 2018.

\bibitem[Xu et~al.(2018)Xu, Yang, Meng, Lu, Huang, Ying, and
  Huang]{xu2018towards}
Zhenbo Xu, Wei Yang, Ajin Meng, Nanxue Lu, Huan Huang, Changchun Ying, and
  Liusheng Huang.
\newblock Towards end-to-end license plate detection and recognition: A large
  dataset and baseline.
\newblock In \emph{Proceedings of the European conference on computer vision
  (ECCV)}, pages 255--271, 2018.

\end{thebibliography}

\end{document}